\documentclass[letterpaper]{article} 
\usepackage[preprint]{aaai27_author_kit/aaai2027} 
\usepackage[hyphens]{url}
\usepackage{graphicx}
\usepackage{natbib}
\usepackage{subcaption}
\usepackage{algorithm}
\usepackage{algorithmic}
\usepackage{booktabs}
\usepackage{amsmath}
\usepackage{amssymb}
\usepackage{float} 
\title{Probabilistic Reachable-Action Verification of Visuomotor Policies \\ via Set-Based Training}
\author{
    Yanliang Huang\textsuperscript{\rm 1},
    Zhuocheng Zhang\textsuperscript{\rm 1},
    Peng Xie\textsuperscript{\rm 1},
    Zhen Zhang\textsuperscript{\rm 1},
    Wenyuan Wu\textsuperscript{\rm 1},
    Majid Khadiv\textsuperscript{\rm 1},
    Zhuoqi Zeng\textsuperscript{\rm 2},
    Amr Alanwar\textsuperscript{\rm 1}
}
\affiliations{
    \textsuperscript{\rm 1}School of Computation, Information and Technology, Technical University of Munich, Munich, Germany\\
    \textsuperscript{\rm 2}School of Engineering, Hainan Bielefeld University of Applied Sciences, Hainan, China\\
    \{yanliang.huang, zhuocheng.zhang, p.xie, zhenzhang.zhang, wenyuan.wu, majid.khadiv, alanwar\}@tum.de,
    zhuoqi.zeng@hainan-biuh.edu.cn
}

\newcommand{\pgdSetMedian}{16.38}
\newcommand{\pgdCtlMedian}{933}
\newcommand{\pgdConsistMedian}{935}
\newcommand{\pgdPGDMedian}{1224}
\newcommand{\pgdControlFloor}{500}
\newcommand{\pgdNCheckpoints}{12}
\newcommand{\pgdNAnchors}{15}
\newcommand{\pgdNEval}{180}

\newcommand{\pgdSetLo}{16.04}
\newcommand{\pgdSetHi}{16.98}

\newcommand{\deepzUsetMedian}{2.1}
\newcommand{\deepzUctlMedian}{1.01e+03}
\newcommand{\deepzUpgdMedian}{743.4}

\newcommand{\deepzViolTrain}{0}
\newcommand{\deepzViolAdapt}{0}

\newcommand{\deepzQUsetMedian}{1.9}
\newcommand{\deepzQUctlMedian}{950.4}
\newcommand{\deepzQUpgdMedian}{686.8}
\newcommand{\deepzBridgeHolds}{all twelve}

\newcommand{\capSmallQ}{0.105}
\newcommand{\capMedQ}{0.060}
\newcommand{\capLargeQ}{0.034}

\begin{document}
\maketitle

\begin{abstract}
Reachability analysis for visuomotor policies is difficult because large visual encoders make end-to-end set propagation computationally expensive and excessively conservative. We therefore freeze the visual encoder and confine set propagation to a low-dimensional interface between it and the downstream policy, with the interface set calibrated from held-out camera-pose perturbations. Propagating this set through the policy with zonotopes yields a terminal output-enclosure width that set-based training optimizes directly. During evaluation, camera-pose perturbations are sampled from the prescribed distribution, and rollout-level split conformal calibration converts the resulting action-deviation scores into a probabilistic reachable-action radius with finite-sample coverage. In controlled manipulation experiments, set-based training reduces this radius while preserving closed-loop task capability, and matched behavior-only, observational-consistency, and pointwise-adversarial controls all leave a larger radius.
\end{abstract}

\section{Introduction}

\begin{figure*}[t]
    \centering
    \includegraphics[width=0.9\textwidth]{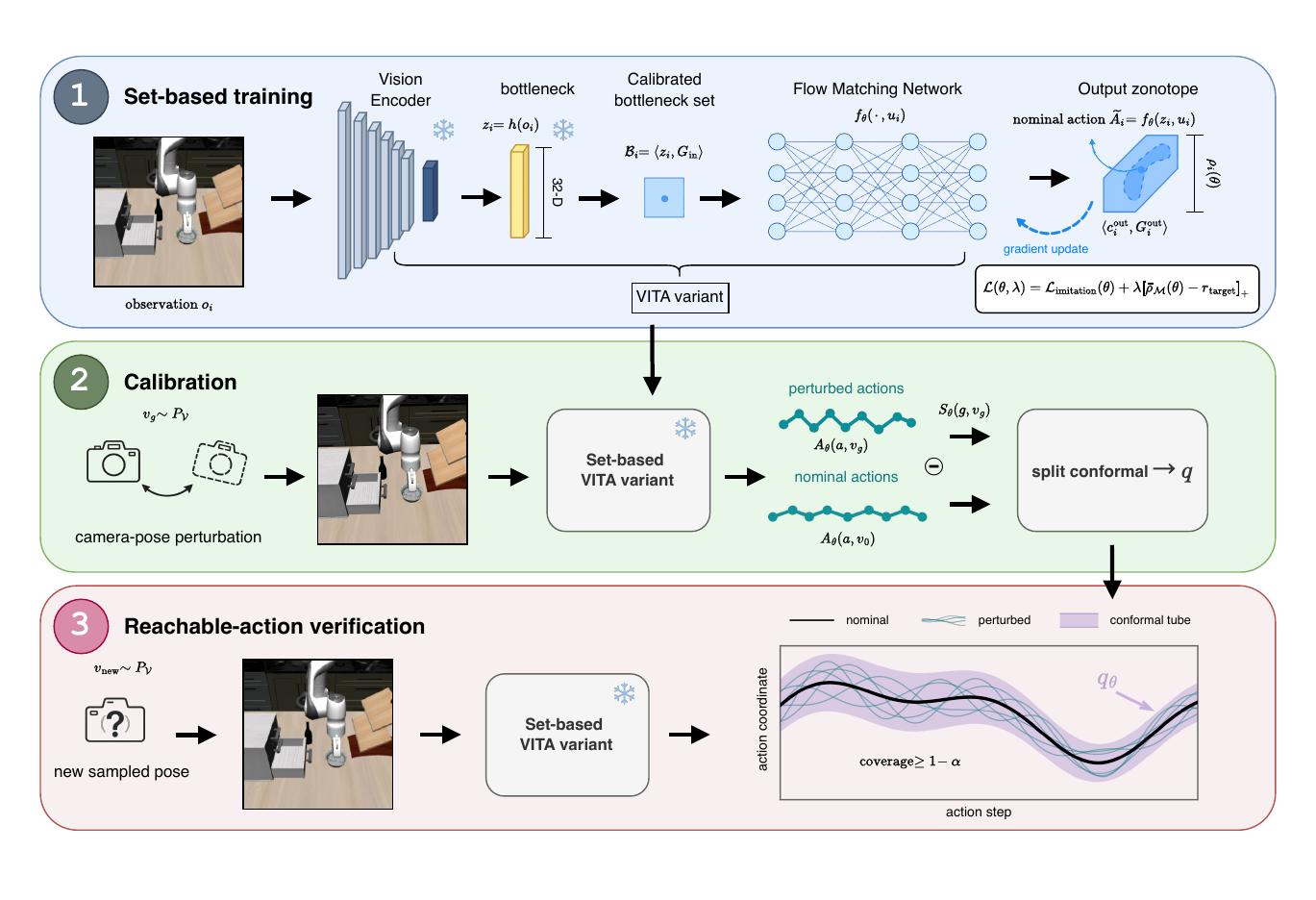}
    \caption{
    Overview of the proposed pipeline.
    \textbf{(1) Set-based training:}
    an observation is mapped through the frozen vision encoder and bottleneck to a representation $z$.
    The calibrated bottleneck set
    $\mathcal{B}_{\epsilon}(z)=\langle z,\epsilon I_m\rangle$
    is propagated through the Vision-to-Action Flow Matching Policy (VITA) \citep{vita} variant to obtain an output zonotope enclosing the actions reachable from that set.
    The objective reduces the enclosure width while preserving imitation performance.
    \textbf{(2) Calibration:}
    the trained policy evaluates nominal and camera-pose-perturbed observations.
    Their rollout-level action deviations are calibrated using split conformal prediction to obtain the probabilistic reachable-action radius $q$.
    \textbf{(3) Probabilistic reachable-action verification:}
    for a new camera pose sampled from the prescribed distribution, $q$ upper-bounds the rollout-level deviation from the nominal action block with finite-sample marginal coverage at least $1-\alpha$.
    }
    \label{fig:pipeline}
\end{figure*}

Fixed cameras on manipulation robots drift through calibration residuals, mounting shifts, and thermal deformation during deployment \citep{leeDreamPose, nobreFastcal, nimuraThermalDrift}. A pose shift changes the observed image and can make a learned visuomotor policy produce a different action sequence for the same scene, and the largest resulting action deviation is the safety concern, since it may drive the end effector into unsafe regions or cause collisions. Recent robustness benchmarks confirm that camera-viewpoint perturbations substantially affect visuomotor policies \citep{liberoPlus}. This motivates verification that quantifies, before deployment, how far a camera-pose perturbation can move the commanded action.

Set-based training makes the size of a propagated output enclosure an explicit training objective, so the trained network produces tighter enclosures that are easier to verify \citep{kollerSetTraining}. Applying it end-to-end to a modern visuomotor policy is intractable, since propagating a set through a large visual encoder accumulates relaxation error across many layers and leaves the resulting action set both expensive to compute and too loose to be useful.

Our idea is to keep the visual encoder frozen and insert a calibrated low-dimensional interface between it and the downstream policy, so that the camera-pose uncertainty is carried in a few dimensions and set propagation runs only through the small downstream policy. Set-based training then contracts the propagated output enclosure directly, which is tractable where end-to-end propagation is not. Split conformal calibration of the sampled action deviations then reports a radius with finite-sample coverage. Figure~\ref{fig:pipeline} summarizes the pipeline.

Our contributions are the following.
\begin{itemize}
\item We decouple the frozen visual encoder from the downstream policy through a calibrated low-dimensional interface and apply set-based training to the policy alone, which makes set-based verification tractable for a visuomotor policy.
\item We prove that the trained enclosure width upper-bounds the physical action deviation induced by a sampled camera perturbation, which links the training objective to the calibrated radius reported at deployment.
\item On a manipulation benchmark under camera-extrinsic perturbation, set-based training reduces the reachable-action radius while preserving closed-loop task capability, and matched behavior-only, observational-consistency, and pointwise-adversarial controls do not reproduce the reduction.
\end{itemize}

\section{Preliminaries and Problem Formulation}\label{sec:problem}

\paragraph{Zonotopes.}
We represent a bounded set by a zonotope~\citep{conf:zono1998}. A zonotope with center $c \in \mathbb{R}^n$ and generator matrix $G \in \mathbb{R}^{n \times p}$ is
\begin{equation}
\mathcal{Z}
=
\langle c,G\rangle
=
\left\{
c+G\beta :
\beta\in[-1,1]^p
\right\}.
\label{eq:zono}
\end{equation}
Its coordinate-wise half-width vector is
\begin{equation}
h(\mathcal{Z}) = |G|\mathbf{1},
\end{equation}
and its $L_\infty$ radius is
\begin{equation}
r_\infty(\mathcal{Z})
=
\|h(\mathcal{Z})\|_\infty
=
\max_j \sum_k |G_{jk}|.
\end{equation}
An affine map $x\mapsto Wx+b$ transforms the zonotope as
\begin{equation}
W\langle c,G\rangle+b
=
\langle Wc+b,WG\rangle.
\label{eq:affine}
\end{equation}
For a ReLU layer we use the single-neuron zonotope relaxation, which appends one generator per crossing coordinate to bound the relaxation error \citep{deepz}. Writing $r_j=\sum_k|G_{jk}|$, $l_j=c_j-r_j$, and $u_j=c_j+r_j$ for the interval induced along coordinate $j$, the relaxation is
\begin{equation}
\mathrm{ReLU}^{\#}\langle c,G\rangle
=
\bigl\langle\, \Lambda c + \mu,\ [\,\Lambda G \mid E\,] \,\bigr\rangle,
\qquad \Lambda=\mathrm{diag}(\lambda_j),
\label{eq:relu}
\end{equation}
with the coordinate-wise slope $\lambda_j$ and offset $\mu_j$ given by
\begin{equation}
\begin{aligned}
\lambda_j &=
\begin{cases}
0, & u_j\le 0,\\[2pt]
1, & l_j\ge 0 \ \text{and}\ u_j>0,\\[2pt]
\dfrac{u_j}{u_j-l_j}, & l_j<0<u_j,
\end{cases}\\[10pt]
\mu_j &=
\begin{cases}
0, & u_j\le 0 \ \text{or}\ l_j\ge 0,\\[6pt]
\dfrac{-\,u_j l_j}{2(u_j-l_j)}, & l_j<0<u_j.
\end{cases}
\end{aligned}
\label{eq:relu_coeffs}
\end{equation}
where $E$ has one column for each crossing coordinate $j$ with $l_j<0<u_j$, equal to $\mu_j$ in row $j$ and zero elsewhere.

\paragraph{Policy and camera-pose uncertainty.}
Let $a$ denote a deployment state, including the scene and task information held fixed during one policy evaluation, and let $v_0$ be the nominal camera pose. A visuomotor policy $\pi_\theta$ produces an action block
\begin{equation}
A_\theta(a,v)\in\mathbb{R}^{T\times d}
\end{equation}
when the state $a$ is observed under camera pose $v$, with $T$ action steps and $d$ action coordinates per step. We write
\begin{equation}
A_\theta(a)=A_\theta(a,v_0)
\end{equation}
for the nominal action block. The admissible camera-pose set is
\begin{equation}
\mathcal{V}
=
\left\{
v :
\|\Delta\mathrm{rot}(v)\|_\infty
\le
\delta_{\mathrm{rot}},
\quad
\|\Delta\mathrm{trans}(v)\|_\infty
\le
\delta_{\mathrm{trans}}
\right\},
\label{eq:camera_set}
\end{equation}
where $\delta_{\mathrm{rot}}$ and $\delta_{\mathrm{trans}}$ specify the rotation and translation budgets. Let $\mathcal{C}\subseteq\{1,\ldots,d\}$ denote the continuous action coordinates used in the deviation metric.

\paragraph{Camera-pose distribution and verification target.}
Let $P_{\mathcal{V}}$ be the prescribed distribution supported on the bounded camera-pose set $\mathcal{V}$, from which deployment poses are drawn. For a pose $v\sim P_{\mathcal{V}}$, define the nominal-centered action deviation
\begin{equation}
d_\theta(a,v)
=
\max_{t,\;j\in\mathcal{C}}
\left|
A_{\theta,tj}(a,v)
-
A_{\theta,tj}(a)
\right|.
\label{eq:deviation}
\end{equation}
The network emits the action block in normalized coordinates as $\tilde{A}_{\theta}$, and the policy commands $A_{\theta,tj}=s_j\tilde{A}_{\theta,tj}+b_j$ with per-coordinate action scale $s_j>0$ and offset $b_j$, so $A_{\theta}$ is in the policy's commanded action units. The offset cancels in the nominal-centered difference, so the deviation~\eqref{eq:deviation} equals $\max_{t,\;j\in\mathcal{C}}s_j\lvert\tilde{A}_{\theta,tj}(a,v)-\tilde{A}_{\theta,tj}(a)\rvert$ and is measured in commanded action units.

\paragraph{Sampling and conformal calibration.}
Let $g$ denote an exchangeable statistical unit, a rollout initialization, containing a finite set of states $\mathcal{A}(g)$. For each unit we draw one camera pose $v_g\sim P_{\mathcal{V}}$, which our experiments instantiate as the uniform law on $\mathcal{V}$, independently across units, and freeze it before any policy output is read. The group score is the largest nominal-centered deviation this single sampled pose induces over the states of the unit,
\begin{equation}
S_\theta(g,v_g)
=
\max_{a\in\mathcal{A}(g)}
d_\theta(a,v_g).
\label{eq:group_score}
\end{equation}
The verification target is
\begin{equation}
q_\theta^{\star}(\alpha)
=
\inf\left\{
q:\;
\Pr_{g,v}\!\left[S_\theta(g,v)\le q\right]
\ge 1-\alpha
\right\},
\label{eq:dist_target}
\end{equation}
the smallest radius the group score respects with probability at least $1-\alpha$ under the joint draw of a rollout initialization and a camera pose $v\sim P_{\mathcal{V}}$. We call $q_\theta^{\star}(\alpha)$ the probabilistic reachable-action radius induced by $P_{\mathcal{V}}$. We estimate $q_\theta^{\star}$ by rollout-level split conformal calibration. Given $n$ exchangeable calibration groups with ordered scores
$S_{(1)}\le\cdots\le S_{(n)}$,
and assuming $k\le n$, split conformal calibration at level $1-\alpha$ uses
\begin{equation}
k
=
\left\lceil
(n+1)(1-\alpha)
\right\rceil,
\qquad
q_\theta
=
S_{(k)}.
\label{eq:conformal_radius}
\end{equation}
For a new exchangeable group $g_{\mathrm{new}}$ with a fresh pose $v_{\mathrm{new}}$ drawn from the same distribution,
\begin{equation}
\Pr\left[
S_\theta(g_{\mathrm{new}},v_{\mathrm{new}})
\le
q_\theta
\right]
\ge
1-\alpha.
\label{eq:conformal_coverage}
\end{equation}
The exchangeable unit is the pair of a rollout initialization and its sampled pose, so $q_\theta$ is the split-conformal estimate of $q_\theta^{\star}$.

\section{Verification-Compatible Set-Based Training}\label{sec:method}

Flow-matching and diffusion policies are typically realized with a large denoising network \citep{diffusionPolicy}, through which set propagation accumulates relaxation error layer by layer. We therefore take the downstream policy $f_\theta$ to be a Vision-to-Action Flow Matching Policy (VITA) \citep{vita} variant, which realizes the action flow with a small multilayer perceptron, so a propagated set traverses few layers and its enclosure stays informative after propagation.
\subsection{Interface and training objective}\label{sec:method-obj}
The verification interface is a fixed low-dimensional bottleneck through which the bounded set enters the downstream policy. Let $h$ map an observation $o_i$ to a bottleneck representation $z_i = h(o_i)\in\mathbb{R}^m$, and let $f_\theta$ be the downstream policy that maps $z_i$, together with any non-visual conditioning $u_i$, to a normalized action block $\tilde{A}_i = f_\theta(z_i,u_i)$. Set-based training updates the downstream parameters $\theta$ while holding the frozen map $h$ fixed. The training objective combines an imitation loss on $\tilde{A}_i$ with a width loss on the propagated set, which replaces the bottleneck representation with an interface set
\begin{equation}
\mathcal{B}_i = \langle z_i, G_{\mathrm{in}}\rangle,
\label{eq:interface_set}
\end{equation}
a zonotope centered at $z_i$ with a fixed generator matrix $G_{\mathrm{in}}$, whose isotropic case $G_{\mathrm{in}}=\epsilon I_m$ appears in Figure~\ref{fig:pipeline}, and propagates it through the same downstream policy. The generator $G_{\mathrm{in}}$ is fixed before training as the $95\%$ held-out state-level marginal quantile of the bottleneck-representation shift that perturbations $v\sim P_{\mathcal{V}}$ induce, calibrated on $180$ held-out states.

\paragraph{Output enclosure and terminal width.}
The exact object the width loss contracts is the image of the interface set under the downstream policy,
\begin{equation}
\mathcal{Y}_i(\theta) = \{\, f_\theta(z,u_i) : z\in\mathcal{B}_i \,\}.
\label{eq:exact_output_set}
\end{equation}
Let $f_\theta^{\#}$ denote the zonotope abstract transformer that propagates $\mathcal{B}_i$ through $f_\theta$ with the affine map~\eqref{eq:affine} and the ReLU relaxation~\eqref{eq:relu} of Section~\ref{sec:problem}, giving an output zonotope
\begin{equation}
\mathcal{Z}^{\mathrm{out}}_i = f_\theta^{\#}(\mathcal{B}_i,u_i) = \langle c^{\mathrm{out}}_i, G^{\mathrm{out}}_i\rangle
\;\supseteq\;
\mathcal{Y}_i(\theta),
\label{eq:abstract_inclusion}
\end{equation}
whose inclusion of $\mathcal{Y}_i(\theta)$ Proposition~1 establishes. All propagated outputs use the normalized coordinates defined in Section~\ref{sec:problem}, where the center $c^{\mathrm{out}}_i$, the generators $G^{\mathrm{out}}_i$, and the normalized nominal action $\tilde{A}_i$ are expressed, and $s_j$ maps each coordinate to its commanded action unit. Let $\mathcal{J}_{\mathrm{train}}$ denote the flattened step--coordinate indices of the $T\times d$ action block used by the objective, all output coordinates including the gripper channel, and $s_j$ denotes the action scale of the coordinate that flattened index $j$ belongs to. The terminal output-enclosure width is
\begin{equation}
\rho_i
\;=\;
\frac{1}{\nu_i}
\max_{j\in\mathcal{J}_{\mathrm{train}}}
s_j
\sum_k
\bigl\lvert G^{\mathrm{out}}_{i,jk}\bigr\rvert,
\label{eq:terminal_width}
\end{equation}
the largest scaled generator row sum over the training coordinates, normalized by $\nu_i = \max(\lVert A_i\rVert_2,\epsilon_0)$, the stop-gradient $L_2$ norm of the commanded nominal action block $A_i$ of Section~\ref{sec:problem}, with a floor $\epsilon_0 = 10^{-9}$. Thus $\rho_i$ is a dimensionless relative enclosure width.

\paragraph{Adaptive width-constrained objective.}
We minimize the mean imitation loss subject to the mean minibatch width $\bar{\rho}_{\mathcal{M}}=\frac{1}{\lvert\mathcal{M}\rvert}\sum_{i\in\mathcal{M}}\rho_i$ not exceeding $r_{\mathrm{target}}$, and optimize this constraint with an adaptive hinge and a projected multiplier while preserving nominal behavior. The training loss couples the imitation objective with a hinge on the excess width,
\begin{equation}
\mathcal{L}(\theta,\lambda)
\;=\;
\mathcal{L}_{\mathrm{imitation}}(\theta)
\;+\;
\lambda
\bigl[\,\bar{\rho}_{\mathcal{M}} - r_{\mathrm{target}}\,\bigr]_+ ,
\label{eq:set_training_loss}
\end{equation}
where $\mathcal{L}_{\mathrm{imitation}}$ is the imitation loss on $\tilde{A}_i$ and $[\,x\,]_+ = \max(0,x)$. The dual variable $\lambda$ follows projected ascent on the constraint violation,
\begin{equation}
\lambda
\;\leftarrow\;
\Pi_{[0,\lambda_{\max}]}
\!\bigl(
\lambda + \eta_\lambda\,(\bar{\rho}_{\mathcal{M}} - r_{\mathrm{target}})
\bigr),
\label{eq:dual_update}
\end{equation}
with step size $\eta_\lambda$ and $\Pi_{[0,\lambda_{\max}]}$ the projection that clamps the multiplier at $\lambda_{\max}$. The downstream policy factors into affine and $\mathrm{ReLU}^{\#}$ layers, namely an initial latent affine map, then $K$ Euler flow steps $x\mapsto x+\mathrm{d}t\,v_{\theta,\tau}(x,u_i)$ with $\mathrm{d}t=1/K$ and a velocity subnetwork of affine and ReLU layers, then a decoder of affine and ReLU layers. Algorithm~\ref{alg:step} summarizes the training step.

\begin{algorithm}[t]
\caption{Set-based training through a verification interface.}
\label{alg:step}
\begin{algorithmic}[1]
\STATE Require: the frozen interface $h$; the downstream architecture $f_\theta$ unrolled into affine and $\mathrm{ReLU}^{\#}$ layers with frozen \textsc{BatchNorm}; the interface generators $G_{\mathrm{in}}$; the action scales $s_j$ and offsets $b_j$ of Section~\ref{sec:problem}; the number of flow steps $K$ with $\mathrm{d}t=1/K$; the target radius $r_{\mathrm{target}}$; the dual step $\eta_\lambda$ and cap $\lambda_{\max}$; the optimizer $\mathrm{Opt}$
\STATE $\lambda \gets 0$
\FOR{each minibatch $\mathcal{M}$}
\FOR{$i \in \mathcal{M}$}
\STATE $z_i \gets h(o_i)$, \ $\tilde{A}_i \gets f_\theta(z_i,u_i)$ \hfill $\triangleright$ imitation loss input
\STATE $A_{i,tj} \gets s_j\tilde{A}_{i,tj}+b_j$ for all $t,j$ \hfill $\triangleright$ commanded units, Section~\ref{sec:problem}
\STATE $\langle c,G\rangle \gets \mathrm{affine}\,\langle z_i, G_{\mathrm{in}}\rangle$ \hfill $\triangleright$ propagated set~\eqref{eq:affine}
\FOR{$\tau = 0,\ \mathrm{d}t,\ \ldots,\ (K-1)\mathrm{d}t$}
\STATE $\langle c_v,G_v\rangle \gets \mathrm{velocity}_{\tau}\,\langle c,G\rangle$ \hfill $\triangleright$ \eqref{eq:affine},~\eqref{eq:relu}
\STATE $\langle c,G\rangle \gets \langle\, c + \mathrm{d}t\,c_v,\ [\,G\mid 0\,] + \mathrm{d}t\,G_v\,\rangle$ \hfill $\triangleright$ \eqref{eq:euler_step}
\ENDFOR
\STATE $\langle c^{\mathrm{out}}_i,G^{\mathrm{out}}_i\rangle \gets \mathrm{decoder}\,\langle c,G\rangle$ \hfill $\triangleright$ $\mathcal{Z}^{\mathrm{out}}_i$~\eqref{eq:affine},~\eqref{eq:relu}
\STATE $\nu_i \gets \max(\lVert A_i\rVert_2,\ \epsilon_0)$ \hfill $\triangleright$ denominator of~\eqref{eq:terminal_width}
\STATE $\rho_i \gets \dfrac{1}{\nu_i}\max_{j\in\mathcal{J}_{\mathrm{train}}} s_j \sum_k \lvert G^{\mathrm{out}}_{i,jk}\rvert$ \hfill $\triangleright$ \eqref{eq:terminal_width}
\ENDFOR
\STATE $\bar{\rho}_{\mathcal{M}} \gets \frac{1}{\lvert\mathcal{M}\rvert}\sum_{i\in\mathcal{M}} \rho_i$
\STATE $\mathcal{L}(\theta,\lambda) \gets \mathcal{L}_{\mathrm{imitation}}(\theta) + \lambda\,[\,\bar{\rho}_{\mathcal{M}} - r_{\mathrm{target}}\,]_+$
\STATE $\theta \gets \mathrm{Opt}(\theta,\, \nabla_\theta \mathcal{L})$ \hfill $\triangleright$ $h$ and bottleneck frozen
\STATE $\lambda \gets \Pi_{[0,\lambda_{\max}]}\!\bigl(\lambda + \eta_\lambda(\bar{\rho}_{\mathcal{M}} - r_{\mathrm{target}})\bigr)$ \hfill $\triangleright$ \eqref{eq:dual_update}
\ENDFOR
\STATE Output: trained downstream parameters $\theta$
\end{algorithmic}
\end{algorithm}

\subsection{Output enclosure and variation bound}\label{sec:method-enclosure}
Each Euler step maps $x\mapsto x+\mathrm{d}t\,v_{\theta,\tau}(x,u_i)$, so the same latent $x$ appears in the increment and as the argument of the velocity network. Treating the two occurrences as independent would enlarge the enclosure, so we zero-pad $G_\tau$ to the columns of $G_v$ and add the two matrices, which keeps one coefficient vector across both.

Proposition 1 (Propagation through the downstream flow encloses the action set).\ Let $\mathcal{Z}_{i,\tau}=\langle c_\tau,G_\tau\rangle$ enclose the exact latent set at step $\tau$, and let $\langle c_v,G_v\rangle$ be the velocity-network zonotope enclosure under the affine map~\eqref{eq:affine} and the relaxation~\eqref{eq:relu}, so that for every $x=c_\tau+G_\tau\beta\in\mathcal{Z}_{i,\tau}$ there is an extension $\beta'$ of $\beta$ with $\lVert\beta'\rVert_\infty\le1$ and $v_{\theta,\tau}(x,u_i)=c_v+G_v\beta'$. Then
\begin{equation}
\mathcal{Z}_{i,\tau+1}=\bigl\langle\, c_\tau+\mathrm{d}t\,c_v,\ [\,G_\tau\mid 0\,]+\mathrm{d}t\,G_v \,\bigr\rangle,
\label{eq:euler_step}
\end{equation}
with $[\,G_\tau\mid 0\,]$ the zero-padding of $G_\tau$ to the columns of $G_v$, encloses every exact update $x+\mathrm{d}t\,v_{\theta,\tau}(x,u_i)$, and inducting through the $K$ steps and the decoder transformer gives $\mathcal{Y}_i(\theta)\subseteq\mathcal{Z}^{\mathrm{out}}_i$.

Proof. For $x=c_\tau+G_\tau\beta\in\mathcal{Z}_{i,\tau}$ the padded matrix satisfies $[\,G_\tau\mid 0\,]\beta'=G_\tau\beta$, and the velocity transformer enclosure gives $v_{\theta,\tau}(x,u_i)=c_v+G_v\beta'$ with $\lVert\beta'\rVert_\infty\le1$. Hence $x+\mathrm{d}t\,v_{\theta,\tau}(x,u_i)=(c_\tau+\mathrm{d}t\,c_v)+([\,G_\tau\mid 0\,]+\mathrm{d}t\,G_v)\beta'$ lies in $\mathcal{Z}_{i,\tau+1}$. The source zonotope $\mathcal{B}_i$ is the base case, and the decoder transformer encloses its input by the same per-layer argument, so the inclusion follows by induction. $\square$

Proposition 2 (Output-variation bound).\ Let $\mathcal{Y}_i(\theta)\subseteq\mathcal{Z}^{\mathrm{out}}_i$ and let $\rho_i$ be the terminal width~\eqref{eq:terminal_width}. The normalized half-range of the exact output set over the training coordinates,
\begin{equation}
D_i(\theta)=\frac{1}{2\nu_i}\sup_{y,y'\in\mathcal{Y}_i(\theta)}\ \max_{j\in\mathcal{J}_{\mathrm{train}}} s_j\,\lvert y_j-y'_j\rvert,
\label{eq:half_range}
\end{equation}
satisfies $D_i(\theta)\le\rho_i$. Since $\tilde{A}_i=f_\theta(z_i,u_i)\in\mathcal{Y}_i(\theta)$, every interface-set output obeys
\begin{equation}
\frac{\max_{j\in\mathcal{J}_{\mathrm{train}}} s_j\,\lvert f_{\theta,j}(z,u_i)-\tilde{A}_{i,j}\rvert}{\nu_i}\le 2\rho_i,\qquad z\in\mathcal{B}_i.
\label{eq:variation_bound}
\end{equation}
Thus minimizing $\rho_i$ contracts an upper bound on the scaled action variation induced by the whole interface set.

Proof. For coordinate $j$ the interval induced by $\mathcal{Z}^{\mathrm{out}}_i$ has half-width $\sum_k\lvert G^{\mathrm{out}}_{i,jk}\rvert$, so any $y,y'\in\mathcal{Z}^{\mathrm{out}}_i\supseteq\mathcal{Y}_i(\theta)$ satisfy $\lvert y_j-y'_j\rvert\le2\sum_k\lvert G^{\mathrm{out}}_{i,jk}\rvert$. Multiplying by $s_j$, taking the maximum over $j$ and the supremum over $\mathcal{Y}_i(\theta)$, and dividing by $2\nu_i$ gives $D_i(\theta)\le\frac{1}{\nu_i}\max_j s_j\sum_k\lvert G^{\mathrm{out}}_{i,jk}\rvert=\rho_i$. Taking $y=f_\theta(z,u_i)$ and $y'=\tilde{A}_i$, both in $\mathcal{Y}_i(\theta)$, gives~\eqref{eq:variation_bound}. $\square$

\subsection{From terminal width to the sampled deviation}\label{sec:method-bridge}
When the perturbed bottleneck representation lies in the propagated interface set, the terminal width upper-bounds the induced physical action deviation, and a bounded-set argument turns this into a conformal statement. For a rollout initialization $g$ and a sampled camera pose $v$, let $z_a^0=h(o(a,v_0))$ and $z_a^v=h(o(a,v))$ denote the nominal and perturbed bottleneck representations for each $a\in\mathcal{A}(g)$, where $o(a,v)$ is the observation at pose $v$. For $r\ge 0$, let $\mathcal{Z}^{\mathrm{out}}_a(r)=\langle c_a^{\mathrm{out}}(r),G_a^{\mathrm{out}}(r)\rangle$ be the output zonotope obtained by propagating the interface set $\langle z_a^0,rI_m\rangle$, and let $\tilde{A}_a=f_\theta(z_a^0,u_a)$ be the normalized nominal action block, in the same normalized coordinates as $c_a^{\mathrm{out}}$ and $G_a^{\mathrm{out}}$, where $u_a$ is the non-visual conditioning of that state. Define
\begin{align}
E_r(g,v)
&=
\Bigl\{
\max_{a\in\mathcal{A}(g)}
\lVert z_a^v-z_a^0\rVert_\infty
\le r
\Bigr\},
\notag\\
U_\theta(g;r)
&=
\max_{\substack{a\in\mathcal{A}(g)\\ t,\,j\in\mathcal{C}}}
s_j\bigl(
\lvert c^{\mathrm{out}}_{a,tj}(r)-\tilde{A}_{a,tj}\rvert
\notag\\
&\qquad\qquad\quad
{}+\textstyle\sum_k\lvert G^{\mathrm{out}}_{a,tjk}(r)\rvert
\bigr).
\label{eq:rollout_width_bound}
\end{align}
Every term is expressed in the raw action units of the score $S_\theta$ through the per-coordinate scale $s_j$, where $t$ indexes the action time steps and $j\in\mathcal{C}$ the continuous coordinates of Section~\ref{sec:problem}, and the center offset $\lvert c^{\mathrm{out}}_{a,tj}(r)-\tilde{A}_{a,tj}\rvert$ accounts for the gap between the propagated center and the nominal action. Writing $U_{\theta,a}(r)=\max_{t,\,j\in\mathcal{C}}s_j\bigl(\lvert c^{\mathrm{out}}_{a,tj}(r)-\tilde{A}_{a,tj}\rvert+\sum_k\lvert G^{\mathrm{out}}_{a,tjk}(r)\rvert\bigr)$ for the per-state bound gives $U_\theta(g;r)=\max_{a\in\mathcal{A}(g)}U_{\theta,a}(r)$.

Proposition 3 (Terminal-width bound for sampled camera perturbations).\ If $E_r(g,v)$ holds, then
\begin{equation}
S_\theta(g,v)\le U_\theta(g;r).
\label{eq:score_width_bridge}
\end{equation}
Consequently, if $\Pr[E_r(g,v)]\ge 1-\gamma$ with $\gamma$ the miss probability of the radius event, then for every $u\ge0$,
\begin{equation}
\Pr\!\left[S_\theta(g,v)>u\right]
\le
\gamma
+
\Pr\!\left[U_\theta(g;r)>u\right].
\label{eq:tail_width_bridge}
\end{equation}
For a sample-specific bound, set $r_a(v)=\lVert z_a^v-z_a^0\rVert_\infty$ and propagate $\langle z_a^0,r_a(v)I_m\rangle$ for each $a$. Define
\begin{equation}
\widetilde U_\theta(g,v)
=
\max_{a\in\mathcal{A}(g)}
U_{\theta,a}\bigl(r_a(v)\bigr).
\label{eq:adaptive_rollout_bound}
\end{equation}
Then $S_\theta(g,v)\le\widetilde U_\theta(g,v)$ for every sampled pair. Given $n$ exchangeable calibration pairs, let $q_\theta^{U}=\widetilde U_{(k)}$ with $k=\lceil(n+1)(1-\alpha)\rceil$. It follows that
\begin{equation}
q_\theta\le q_\theta^{U},
\qquad
\Pr\!\left[
S_\theta(g_{\mathrm{new}},v_{\mathrm{new}})
\le q_\theta^{U}
\right]
\ge 1-\alpha.
\label{eq:conformal_width_bridge}
\end{equation}

The result follows from interface-set containment, Proposition~1, the triangle inequality with the retained center offset, and order-statistic monotonicity. The proof is given in the supplement.

The chain $\rho\to U_\theta\to q_\theta^U$ holds under two conditions: $q_\theta$ and $q_\theta^U$ are calibrated on the same rollout-pose pairs at the same rank $k$, and the coordinate set $\mathcal{C}$ is the continuous non-gripper set the score $S_\theta$ uses. The bound is a surrogate, and the directly calibrated radius $q_\theta$ stays the tighter primary result.

\section{Experiments}\label{sec:experiments}

\subsection{Experimental Setup}\label{sec:setup}

We evaluate the method on the fixed-camera LIBERO-10 manipulation benchmark under camera-extrinsic perturbations \citep{liberoBenchmark}. The primary controlled study uses task~3, ``put the black bowl in the bottom drawer of the cabinet and close it.'' All four training configurations are available on it under the same pretrained checkpoint, data, fine-tuning budget, and evaluation protocol, so only the training objective differs. The four configurations are behavior-only fine-tuning, observational consistency, PGD adversarial training \citep{madry}, and the proposed set-based objective. Behavior-only fine-tuning holds the width multiplier $\lambda$ of Equation~\eqref{eq:set_training_loss} at zero, observational consistency aligns predictions across the clean and perturbed views, and PGD adversarial training replaces the propagated width with the largest deviation a projected-gradient attack finds inside the same interface set. Each configuration is trained with three random seeds, labeled s621, s622, and s623. We additionally apply the baseline-versus-set-based protocol to all ten LIBERO-10 tasks.

\paragraph{Downstream policy architecture.} All four training configurations use the same verification-compatible VITA variant. Two components of the released architecture are replaced so that every downstream layer is an affine map or a ReLU and the zonotope transformers of Section~\ref{sec:problem} cover the propagated network. Evaluation-mode BatchNorm folds exactly into an affine map, whereas LayerNorm is input-dependent and cannot be folded into a fixed affine transformer. The ReLU relaxation~\eqref{eq:relu} is exact on every sign-stable interval, whereas smooth activations such as GELU admit relaxations but none that is exact, so every unit contributes relaxation error at every layer. We therefore replace the GELU activations with ReLU and the LayerNorm layers with BatchNorm, so relaxation error is paid only at crossing coordinates. A frozen ImageNet-pretrained ResNet18 encoder feeds a learned $32$-dimensional bottleneck, and the downstream network that set-based training updates is an affine map, a single Euler flow step with $K=1$, and an affine--ReLU decoder producing the $10\times7$ action block. The full interface architecture is in the supplement.

\paragraph{Held-out evaluation set.} All task~3 models are evaluated on held-out rollout initializations under the camera-extrinsic perturbation, none of which enters checkpoint selection or pose tuning, and at each evaluation state only the camera pose changes while the robot and scene stay fixed. The formal comparison of Table~\ref{tab:formal} and the augmentation study of Figure~\ref{fig:results-a} calibrate the radius on $200$ initializations and report the empirical coverage on $100$ held-out test initializations, with each initialization scored by the largest sampled deviation over its states. The set $\mathcal{A}(g)$ of each initialization holds $16$ states, collected once by replaying two fixed reference-policy clean rollouts from that initialization, an augmentation-only reference checkpoint and a set-based reference checkpoint contributing eight replan-step states each. These states are stored offline as simulator states with their nominal camera parameters, so they are fixed before evaluation and shared identically across the four training configurations, and every configuration is scored on the same states under the same single sampled pose per initialization.

\paragraph{Camera-pose sampling.} For each rollout initialization we draw one camera pose from the six-dimensional perturbation box, each coordinate independent and uniform over its budget, and freeze it before any model output is read. The same pose applies to every evaluation state of that initialization and to every training configuration at that seed; the three seeds are independent replications that each draw their own poses, and calibration and test initializations draw disjoint poses.

\paragraph{Metrics.} The primary result is the rollout-level conformal radius $q$. Using $n=200$ calibration initializations $q$ is computed at the rank defined in Equation~\eqref{eq:conformal_radius}, reported at coverage levels $1-\alpha\in\{0.90,0.95,0.99\}$. A configuration is behavior-valid when its closed-loop success is at least $0.85$ under both clean and camera-extrinsic-perturbed conditions, which excludes action collapse. The radius uses only the continuous translation and rotation coordinates, since the gripper command is a near-discrete switch, and the gripper channel is reported separately. In the ten-task protocol each of the two compared methods is evaluated with $50$ clean and $50$ perturbed rollouts under matched initial states and a pre-specified episode seed.

Further protocol details are in the supplement.

\subsection{Calibrated Radius Reduction}

Set-based training reduces the conformal radius while keeping the policy behavior-valid, as Table~\ref{tab:formal} shows. At the $99\%$ coverage level it gives a median per-seed ratio of $2.09\times$ over behavior-only fine-tuning, with per-seed radii in Table~\ref{tab:formal}, while observational consistency stays at the behavior-only level. The empirical coverage on the $100$ held-out initializations is $0.99$ to $1.00$. Behavior-only fine-tuning, observational consistency, and set-based training pass the behavior criterion on all three seeds, PGD adversarial training passes the criterion on two of three seeds, and set-based training is the tightest valid configuration on every seed. The same per-seed ordering holds at the $90\%$, $95\%$, and $99\%$ coverage levels reported in the supplement.

\begin{table}[t]
\centering
\small
\setlength{\tabcolsep}{5pt}
\caption{Rollout-initialization-level $99\%$ conformal reachable-action radius $q$ under bounded sampled six-dimensional camera-extrinsic perturbation, on $200$ calibration and $100$ held-out test initializations, reported per seed with the three-seed median. Each initialization draws one frozen camera pose that is shared across all four configurations at that seed, which differ only in the training signal. Empirical coverage on the held-out initializations is $0.99$ to $1.00$. Set-based training is the tightest configuration on every seed, a median per-seed ratio of $2.09\times$ over behavior-only fine-tuning. The dagger marks the seed-621 PGD adversarial-training checkpoint, which fails the closed-loop behavior gate and is reported as a frontier point, excluded from the behavior-matched controls. Lower is better.}
\label{tab:formal}
\begin{tabular}{lcccc}
\toprule
Training signal & $q_{\mathrm{s621}}$ & $q_{\mathrm{s622}}$ & $q_{\mathrm{s623}}$ & median \\
\midrule
Behavior-only fine-tuning & 0.231 & 0.181 & 0.191 & 0.191 \\
Observational consistency & 0.220 & 0.201 & 0.186 & 0.201 \\
PGD adversarial training & 0.173$^{\dagger}$ & 0.138 & 0.114 & 0.138 \\
\textbf{Set-based training} & \textbf{0.111} & \textbf{0.113} & \textbf{0.077} & \textbf{0.111} \\
\bottomrule
\end{tabular}
\end{table}

\begin{figure}[t]
\centering
\includegraphics[width=0.6\linewidth]{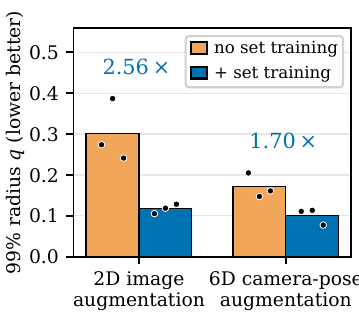}
\caption{Rollout-initialization-level $99\%$ conformal reachable-action radius $q$ on the held-out Task~3 evaluation set, comparing two augmentation protocols, two-dimensional image jitter and six-dimensional camera-extrinsic perturbation, each without and with set-based training. Bars are three-seed means with individual seeds as dots, lower is better. The annotated factors are the set-versus-no-set reductions of three-seed means within each augmentation protocol, and the three-seed mean radii of each bar are given in the text.}
\label{fig:results-a}
\end{figure}

The reduction remains when the augmentation matches the physical threat, as Figure~\ref{fig:results-a} summarizes, where the baselines are augmentation-only policies trained separately from the matched continuation configurations of Table~\ref{tab:formal}. On the rollout-initialization-level $99\%$ radius, set-based training tightens the three-seed mean from $0.301$ to $0.117$ under two-dimensional augmentation, a $2.56\times$ reduction, and from $0.171$ to $0.101$ under six-dimensional camera-extrinsic augmentation, a $1.70\times$ reduction, with all three six-dimensional set-based seeds behavior-valid. Because the reduction persists under the six-dimensional augmentation matched to the threat model, the tightening reflects the training objective.

\subsection{Training Signal and Enclosure Tightness}

Under the matched protocol of Section~\ref{sec:setup}, observational consistency constrains paired outputs yet leaves the calibrated radius at the behavior-only level. PGD adversarial training reduces the radius on all three seeds but is behavior-valid on only two, and its seed-621 checkpoint is reported only as a frontier point. The set-based radius is smaller than the PGD radius on all three seeds. On the two behavior-valid seeds the paired ratios $q_{\mathrm{PGD}}/q_{\mathrm{set}}$ are $1.22$ and $1.48$, and their $95\%$ paired bootstrap intervals exclude one, so the behavior-matched comparison uses those two seeds. The two objectives differ mechanistically: PGD optimizes the largest deviation a projected-gradient attack locates, whereas set-based training optimizes an over-approximation of the entire interface-set output. This distinction shows in the deterministic enclosure. The ratio of the zonotope width to the strongest attack inside the same set has a median of \pgdSetMedian{} for set-based training and exceeds \pgdControlFloor{} for the three controls, so only set-based training keeps this enclosure tighter and more informative under the identical verifier, as the supplement details. This enclosure is the training surrogate for the whole interface-set output on the fixed bottleneck, while the deployment guarantee comes from the rollout-level conformal radius, which Proposition~3 connects to the trained width.

We next evaluate the bound of Proposition~3 on the same evaluation set. The fixed training radius covers $0.76$ to $1.00$ of the test groups across seeds, reflecting the independently drawn poses of the three seed replications, and no covered group violates the fixed-radius bound. The sample-adaptive bound holds on every evaluated group, so $q_\theta\le q_\theta^{U}$ on \deepzBridgeHolds{} checkpoints, and set-based training reduces the median adaptive conformal bound $q_\theta^{U}$ by more than two orders of magnitude relative to every control, as the supplement details. This bound links the trained width to the physical score, while the directly calibrated radius $q_\theta$ remains the tighter primary result.

\subsection{Cross-Task Results}

\begin{figure}[t]
\centering
\includegraphics[width=0.8\linewidth]{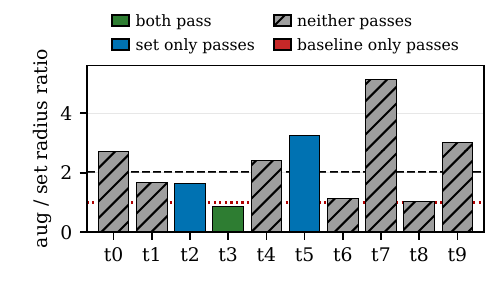}
\caption{Ten-task breadth of set-based training. The per-task ratio of the two-dimensional image-augmentation radius to the set-based radius across all ten tasks under the common two-dimensional image augmentation, colored by which methods pass the behavior criterion at the $0.85$ gate, where green marks the tasks both methods pass, blue marks the tasks only set-based training passes, hatched gray marks the tasks neither method passes, and red marks the tasks only the baseline passes. The dashed line marks the all-task median and the dotted line marks a ratio of one. The task instruction for each index is listed in the supplement.}
\label{fig:results-b}
\end{figure}

The effect extends across tasks with a task-dependent reach. Under a common two-dimensional image augmentation applied to all ten tasks, the median ratio of the augmentation-only radius to the set-based radius is $2.03$ over the ten tasks, as Figure~\ref{fig:results-b} shows. To separate weak baseline policies from training-induced behavior change, we apply the behavior criterion of Section~\ref{sec:setup} as a minimum task-capability requirement, under a shared rule that selects each method's best validation-loss checkpoint with matched initial states and a pre-specified episode seed. At this criterion the augmentation baseline is capable on the one task, task $3$, whereas set-based training is behavior-valid on three tasks, tasks $2$, $3$, and $5$, so it preserves the one baseline-capable task and recovers two further tasks that the augmentation baseline does not reach, with no task passing only under the baseline. Among the three tasks where set-based training is behavior-valid, it is tighter on tasks $2$ and $5$ and slightly looser on task $3$, yielding a median augmentation-to-set radius ratio of $1.65$ under this separate single-seed breadth protocol. The all-task median of $2.03$ is descriptive and includes behavior-invalid tasks. The classification and its sensitivity to the criterion are reported in the supplement.

A capacity study across three widths of the flow matching network is reported in the supplement. Direct contraction at the raw 1024-dimensional feature interface attains small training widths but does not pass the behavior criterion, as the supplement reports, which supports the compact bottleneck as the behavior-compatible interface.

\section{Related Work}

\paragraph{Visual robustness under camera and observation shifts.}
Camera-motion smoothing and pixel-wise smoothing provide certified robustness for visual perception under structured camera or image perturbations, while viewpoint adversarial search studies the sensitivity of recognition systems to changes in three-dimensional viewpoint \citep{camMotionSmoothing,pixelwiseSmoothing,viewfool}. Reachability-based pose estimation provides guarantees for a perception front-end by bounding the estimated camera pose from visual observations \citep{ladnerPoseCert}. In visuomotor learning, augmentation exposes policies to sampled visual transformations \citep{diffusionPolicy}, and observational-consistency objectives regularize action predictions across perturbed observations \citep{rovla}. We extend this line to action-space verification.

\paragraph{Set-based neural verification and training.}
Abstract interpretation, zonotope domains, and convex or linear relaxations propagate bounded inputs through neural networks to construct output enclosures \citep{deepz,deeppoly,wongKolter,crown}. Differentiable verification bounds have further been incorporated into certified training objectives \citep{diffai,wongKolter,autoLirpa}. Set-based training directly optimizes the size of the propagated output enclosure, improving the robustness and subsequent verifiability of the trained network \citep{kollerSetTraining}. Related work extends set propagation to reinforcement-learning policies, to latent spaces of learned encoders, and to neural barrier and Lyapunov certificates \citep{wendlSetRL,pointsToSets,kranzlmullerBarrier,certifiedSetConv}. We apply this line to a visuomotor policy by confining propagation to a calibrated low-dimensional interface, which keeps the enclosure informative where end-to-end propagation is not.

\paragraph{Conformal uncertainty quantification for control.}
Conformal prediction constructs data-dependent prediction sets and thresholds with finite-sample marginal coverage and has been applied to learned control, reachability, and disturbance uncertainty in robotics \citep{conformalDagger,conformalOSC,cddrReach}. Zono-conformal prediction further studies structured multi-output prediction sets and the trade-off between coverage and conservatism \citep{zonoConformal}. We apply split conformal calibration \citep{leiDistfree} to rollout-level action-deviation scores, giving a probabilistic reachable-action radius for a visuomotor policy.

\section{Conclusion}

Our study shows that the geometry of a policy's action set under sensor uncertainty can be shaped during training once perception is decoupled from downstream set reasoning through a calibrated low-dimensional interface, which makes a full-set training signal affordable. The distinction from pointwise adversarial training is mechanistic: the set objective constrains the entire enclosure, whereas a pointwise attack constrains only the point it locates. This mechanism is consistent with the smaller physical action deviations observed for set-based training.

A deterministic map from the pose set to the interface set would upgrade the probabilistic radius toward an end-to-end guarantee.

\bibliography{references}

\appendix
\setcounter{figure}{0}
\setcounter{table}{0}
\setcounter{equation}{0}
\setcounter{algorithm}{0}
\renewcommand{\thefigure}{S\arabic{figure}}
\renewcommand{\thetable}{S\arabic{table}}
\renewcommand{\theequation}{S\arabic{equation}}
\renewcommand{\thealgorithm}{S\arabic{algorithm}}
\makeatletter
\setlength{\@fptop}{0pt}
\setlength{\@fpsep}{12pt plus 0fil}
\setlength{\@fpbot}{0pt plus 1fil}
\makeatother
\section{Bounded Camera-Perturbation Sampling and Conformal Rank}
The evaluation perturbs the single fixed agentview camera inside a bounded six-dimensional box. The three rotation coordinates are the world-frame roll, pitch, and yaw in degrees, and the three translation coordinates are the camera-centre displacements in millimetres, so a pose offset is $v=(\mathrm{roll},\mathrm{pitch},\mathrm{yaw},\mathrm{d}x,\mathrm{d}y,\mathrm{d}z)$. Each coordinate is drawn independently and uniformly, the rotations over $[-1,1]$ degrees and the translations over $[-1,1]$ millimetres. For each rollout initialization $g$ we draw one offset $v_g$ from this distribution under a fixed seed, freeze it before any model output is read, and apply the same $v_g$ to every evaluation state of $g$ and to every checkpoint and training configuration. Calibration and test initializations draw disjoint independent offsets. The rollout-level score is the largest non-gripper action deviation induced by the frozen pose, $S_\theta(g,v_g)=\max_{a}\max_{t,j} |A_{\theta,tj}(a,v_g)-A_{\theta,tj}(a,v_0)|$ over the states $a$ of $g$, the prefix steps $t$, and the non-gripper action coordinates $j$, and the gripper channel is recorded separately. The conformal radius is the order statistic $S_{(k)}$ with $k = \lceil (n+1)(1-\alpha) \rceil$ over the calibration groups, and all boundary ties at the rank are retained. The pose manifest that fixes every $v_g$, together with the calibration and test initialization indices, is released with the code.
\section{Bottleneck-Set Training Radius and Architecture}
\paragraph{Verification interface architecture.} The verification interface is a fixed 32-dimensional linear bottleneck between a frozen visual encoder and the downstream policy. Each of the two fixed views, the agentview and the wrist eye-in-hand, passes through a frozen ImageNet-pretrained ResNet18 with frozen batch normalization, and the two 512-dimensional penultimate features concatenate into the 1024-dimensional visual feature. The bottleneck is a single linear map from 1024 to 32 dimensions with a bias term, no activation, and no output normalization, so the bottleneck representation is an affine image of the frozen feature, and the 32-dimensional representation then joins the proprioceptive state and the task embedding before the linear observation encoder. Its 32-by-1024 weight is initialized from the singular value decomposition of the base checkpoint feature-to-representation projection, and the dimension 32 is chosen from a 16, 32, 64 sweep in which 32 retains the best clean and image-jitter behavior. The base policy is trained first and the bottleneck is inserted afterward, so it is a post-hoc compression of an already-trained feature. The bottleneck is then trained in a first stage under image augmentation of plus or minus five degrees and plus or minus eight pixels, unfreezing only the bottleneck and the encoder while the flow and decoder stay frozen, per task, under the same behavior objective as the base policy, which combines a flow-matching term with action-encoder and flow-decoder reconstruction and a latent-consistency term, so the bottleneck reduces the propagation dimension from 1024 to 32 and is trained to reduce representation shifts induced by camera perturbations. The propagated downstream policy uses ReLU activations and batch normalization, replacing the smooth-activation default, so the set propagation is exact except for the single-neuron ReLU relaxation. In the contraction stage the encoder, the bottleneck, and the two BatchNorm layers are all frozen and only the downstream flow and decoder parameters are updated, so the interface-set calibration remains valid, and the frozen BatchNorm layers run in evaluation mode in the first stage and in contraction alike, with their statistics folded into the affine maps for propagation. Every Task-3 configuration in the main comparison shares this same frozen bottleneck.

\paragraph{Downstream policy architecture.} The compact policy is a flow-matching policy. The downstream policy receives the $32$-dimensional bottleneck representation together with an $8$-dimensional proprioceptive state and a $16$-dimensional task embedding. The downstream policy that set-based training updates and that zonotope propagation traverses is an initial latent affine map into a $512$-dimensional latent, then $K$ Euler flow steps whose velocity subnetwork is a feedforward stack of affine and ReLU layers with frozen BatchNorm folded into the affine maps, then an affine--ReLU decoder that outputs the $10\times7$ action block, namely $T=10$ steps by $d=7$ coordinates. The small, medium, and large capacity tiers set the velocity subnetwork to two layers of width $512$, four layers of width $512$, and four layers of width $1024$, over the same frozen bottleneck and decoder. The reported policies use a single Euler step, so $K=1$ and $\mathrm{d}t=1$. The velocity subnetwork and the decoder both use ReLU activations and BatchNorm.

The set-based training input radius $\epsilon$ is the half-width of the isotropic $L_\infty$ box $\mathcal{B}_\epsilon(z)$ that the objective perturbs around the clean standardized bottleneck representation $z$, carried as a zonotope. Both set-based configurations propagate the same calibrated diagonal input set of half-width $0.240$ in standardized bottleneck-representation units. This training radius is itself the bottleneck-residual calibration. For each state the clean observation and a $\pm1$ degree, $\pm1$ millimeter perturbed observation pass through the frozen encoder and the learned 32-dimensional bottleneck, and the $L_\infty$ change in the representation over 180 states, split into 120 calibration and 60 test, has a split-conformal 95th percentile of $0.240$ in bottleneck-representation units with the held-out test residual at the same scale. The calibration initializations are the held-out extrinsic-residual states 35 through 49, fixed before contraction training and disjoint from the checkpoint-selection states on initializations 0 through 17, so the training radius is a held-out residual quantile computed from data. The shared radius $\epsilon=0.240$ is a held-out state-level marginal residual quantile that defines the training interface, and its rollout-initialization-level coverage is reported separately in the Proposition~3 empirical support.

\section{Held-Out Evaluation Set and Initialization-Level Coverage}
The held-out Task~3 evaluation set of the main text is the rollout-initialization population of the sampled-camera protocol. The rollout initialization is the conformal unit, and each initialization is scored by the worst sampled deviation over its states under its frozen camera pose. On $200$ calibration and $100$ held-out test initializations, split conformal at rank $k=\lceil (200+1)(1-\alpha) \rceil$ gives the per-seed radii of the main-paper table at the $99\%$ level, and Table~\ref{tab:formal-levels} lists all three levels. Set-based training is the tightest configuration on every seed at every level, a median $2.09\times$ tightening over behavior-only fine-tuning at the $99\%$ level, and the empirical coverage on the $100$ held-out initializations is $0.99$ to $1.00$. The set-versus-control ordering and the near-twofold tightening are unchanged across the $90\%$, $95\%$, and $99\%$ levels.
\begin{table}[t]
\centering
\small
\setlength{\tabcolsep}{5pt}
\caption{Rollout-initialization-level conformal reachable-action radius $q$ at the $90\%$, $95\%$, and $99\%$ coverage levels, per seed, on $200$ calibration and $100$ held-out test initializations under bounded sampled camera perturbation. The set-versus-control ordering and the tightening hold at every level, so the main-paper $99\%$ result is not an artifact of a single coverage choice.}
\label{tab:formal-levels}
\begin{tabular}{ll ccc}
\toprule
Training signal & seed & $q_{90}$ & $q_{95}$ & $q_{99}$ \\
\midrule
Behavior-only fine-tuning & s621 & 0.126 & 0.150 & 0.231 \\
                           & s622 & 0.133 & 0.149 & 0.181 \\
                           & s623 & 0.119 & 0.143 & 0.191 \\
\addlinespace
Observational consistency & s621 & 0.125 & 0.147 & 0.220 \\
                           & s622 & 0.137 & 0.159 & 0.201 \\
                           & s623 & 0.118 & 0.137 & 0.186 \\
\addlinespace
PGD adversarial training & s621 & 0.108 & 0.121 & 0.173 \\
                           & s622 & 0.087 & 0.107 & 0.138 \\
                           & s623 & 0.081 & 0.088 & 0.114 \\
\addlinespace
Set-based training & s621 & 0.075 & 0.085 & 0.111 \\
                           & s622 & 0.077 & 0.089 & 0.113 \\
                           & s623 & 0.053 & 0.061 & 0.077 \\
\bottomrule
\end{tabular}
\end{table}

Algorithm~\ref{alg:calibration} states this rollout-level calibration in pseudocode.

\begin{algorithm}[t]
\caption{Rollout-level conformal calibration under bounded camera-perturbation sampling.}
\label{alg:calibration}
\begin{algorithmic}[1]
\STATE \textbf{Require:} frozen policy $\pi_\theta$; bounded camera-pose distribution $P_{\mathcal{V}}$; an initialization-grouped calibration and test split $\mathcal{G}_{\mathrm{cal}},\mathcal{G}_{\mathrm{test}}$ with group contexts $\mathcal{A}(g)$; level $\alpha$
\FOR{each group $g \in \mathcal{G}_{\mathrm{cal}} \cup \mathcal{G}_{\mathrm{test}}$}
\STATE draw $v_g \sim P_{\mathcal{V}}$ and freeze it \hfill $\triangleright$ one pose per initialization, before any model output
\FOR{each context $a \in \mathcal{A}(g)$}
\STATE $A_\theta(a) \gets A_\theta(a,v_0)$ \hfill $\triangleright$ nominal action block
\ENDFOR
\STATE $S_\theta(g) \gets \max_{a \in \mathcal{A}(g)} d_\theta(a,v_g)$ \hfill $\triangleright$ deviation at the frozen pose, group score (statistical unit)
\ENDFOR
\STATE $k \gets \lceil (|\mathcal{G}_{\mathrm{cal}}| + 1)(1 - \alpha) \rceil$
\STATE $q_\theta \gets k\text{-th smallest of } \{S_\theta(g) : g \in \mathcal{G}_{\mathrm{cal}}\}$ \hfill $\triangleright$ ties kept
\STATE $\widehat{\mathrm{cov}} \gets |\{g \in \mathcal{G}_{\mathrm{test}} : S_\theta(g) \le q_\theta\}| \,/\, |\mathcal{G}_{\mathrm{test}}|$ \hfill $\triangleright$ coverage check
\STATE \textbf{Output:} radius $q_\theta$; empirical coverage $\widehat{\mathrm{cov}}$
\STATE \textbf{Guarantee:} a new exchangeable group $g_{\mathrm{new}}$ with a fresh pose $v_{\mathrm{new}} \sim P_{\mathcal{V}}$ has $S_\theta(g_{\mathrm{new}},v_{\mathrm{new}}) \le q_\theta$ with probability at least $1-\alpha$
\end{algorithmic}
\end{algorithm}

\section{Shared Interface Set and Behavior-Valid Selection}
\paragraph{The fixed shared input set.}
All three seeds of the six-dimensional set-based configuration are trained using the same fixed input set, the isotropic bottleneck box $0.240\,I_{32}$ of half-width $0.240$ in standardized bottleneck-representation units. This box is a single training-time input set that is identical across the three seeds and is larger and more conservative than any per-seed radius fit to each seed's pretrained checkpoint, so every seed is contracted against the same fixed conservative choice. With this single shared box, all three seeds reach the 0.85 behavior gate.

\paragraph{Seed-specific recalibration sensitivity.}
As a sensitivity variant, we instead recalibrate the contraction box per seed from each seed's pretrained checkpoint bottleneck residual. This seed-specific box is smaller than the shared box on some seeds and causes one seed to fall below the behavior gate, so the seed-specific variant is behavior-valid on two of three seeds. The larger shared box therefore gives better behavior than the tighter seed-specific box, which we record as a non-monotone training result. The reported six-dimensional set-based configuration uses the shared box, and the seed-specific variant is this sensitivity ablation.

\paragraph{Observational-consistency configuration and a two-dimensional-affine ablation.}
The observational-consistency configuration of the main-paper table is trained under the same six-dimensional camera-extrinsic threat as the set-based configuration. It shares the pretrained checkpoint, the six-dimensional camera-extrinsic perturbation cache, the training budget, the checkpoint schedule, and the behavior-gated selection with the set-based configuration, and it differs only in the training objective, which is the RoVLA and Pi-model consistency penalty at the fixed weight $0.3$ with the set-based width loss disabled and frozen BatchNorm and bottleneck. All three seeds pass the behavior gate, at clean and camera-extrinsic success $0.88$ and $0.94$, $0.96$ and $0.92$, and $0.94$ and $0.94$, and their per-seed $q_{90}$, $q_{95}$, and $q_{99}$ are the observational-consistency rows of Table~\ref{tab:formal-levels}. Under this matched six-dimensional threat the consistency radius stays at the behavior-only level, while set-based training contracts it. The consistency penalty pulls the clean-view and perturbed-view actions together without directly contracting the full interface-set output enclosure, whereas the set-based objective contracts that enclosure. Under the matched threat, only the set-based objective produces the observed tightening. A separate two-dimensional-affine consistency control, trained under in-plane image augmentation, is behavior-valid on only one of three seeds.

\paragraph{PGD adversarial-training configuration.}
The PGD adversarial-training configuration shares the pretrained checkpoint, the six-dimensional camera-extrinsic cache, the training budget, the checkpoint schedule, and the behavior-gated selection with the set-based configuration under the identical adaptive Lagrangian, and it differs only in the penalty, which replaces the zonotope terminal width with the largest non-gripper action deviation that a projected-gradient attack attains inside the same interface set $\mathcal{B}_\epsilon(z)=\langle z,\epsilon I_{32}\rangle$. The inner attack runs forty sign-gradient ascent steps from four random restarts, with step size $2.5/40$ in the normalized generator coordinates $\alpha\in[-1,1]^{32}$ and each step projected back onto the interface box, and the penalized deviation is normalized by the commanded nominal action norm $\nu_i$ exactly as the terminal width is. This training-time inner attack is distinct from the enclosure-tightness evaluation attack, which runs fifty restarts of one hundred steps.

\paragraph{Checkpoint schedule selection.}
Each seed exposes a fixed epoch schedule over epochs 40, 80, 120, 160, and 200, and for every schedule entry we record its training terminal width, its dual weight, and its clean and camera-extrinsic behavior gate. The selected checkpoint is the behavior-valid schedule entry with the smallest training terminal width, and no best-loss checkpoint is used. The reported six-dimensional set-based radii follow from this selection under the shared fixed input set. The behavior gate uses the same fifty clean and fifty camera-extrinsic rollout episodes over initializations 0 through 49 that the paper reports for each configuration. Because these episodes are also used for checkpoint selection, the reported behavior success is not an independent held-out estimate. A separate re-evaluation of the set-based configuration, on fifty clean and fifty camera-extrinsic rollouts drawn from a disjoint three-hundred-initialization pool with the Task-3 checkpoints fixed and fresh episode randomness, keeps every set-based seed behavior-valid, with held-out clean and camera-extrinsic success $0.88/0.92$, $0.88/0.98$, and $0.86/0.94$ for the three seeds against the checkpoint-selection estimates $0.94/0.94$, $0.88/0.92$, and $0.88/0.86$, each difference within the binomial half-width of the fifty-rollout estimate. The reported radius $q$ is separate, since the behavior gate and the radius selection use held-out initializations disjoint from the calibration and test sets, so $q$ remains an independent held-out result.

\section{Ten-Task Breadth Table}
Table~\ref{tab:libero10} lists the exact benchmark language instruction and the short figure name for each of the ten tasks, read from the LIBERO-10 task map in the default benchmark order.
\begin{table}[t]
\centering
\small
\setlength{\tabcolsep}{4pt}
\caption{The ten LIBERO-10 tasks with their exact benchmark language instructions and the short names used in the figures. Task indices are counted from zero in the default benchmark order. Task~3 is the primary controlled study.}
\label{tab:libero10}
\begin{tabular}{@{}c l p{5.0cm}@{}}
\toprule
task & short name & language instruction \\
\midrule
t0 & soup + sauce & put both the alphabet soup and the tomato sauce in the basket \\
t1 & cheese + butter & put both the cream cheese box and the butter in the basket \\
t2 & stove + moka pot & turn on the stove and put the moka pot on it \\
t3 & bowl in drawer & put the black bowl in the bottom drawer of the cabinet and close it \\
t4 & two mugs & put the white mug on the left plate and put the yellow and white mug on the right plate \\
t5 & book in caddy & pick up the book and place it in the back compartment of the caddy \\
t6 & mug + pudding & put the white mug on the plate and put the chocolate pudding to the right of the plate \\
t7 & soup + cheese & put both the alphabet soup and the cream cheese box in the basket \\
t8 & two moka pots & put both moka pots on the stove \\
t9 & mug in microwave & put the yellow and white mug in the microwave and close it \\
\bottomrule
\end{tabular}
\end{table}

\begin{table}[t]
\centering
\footnotesize
\setlength{\tabcolsep}{3pt}
\caption{Ten-task breadth over held-out initial states under the fixed two-dimensional reference augmentation protocol. The clean and camera-extrinsic columns are matched $50+50$ closed-loop rollout success rates for the augmentation-only baseline and for set-based training. The $q$ columns are the rollout-initialization-level $99\%$ sampled conformal radius and are unchanged by the criterion. A method satisfies the minimum task-capability criterion when both its clean and camera-extrinsic closed-loop success rates are at least $0.85$, and the same criterion is applied to both methods. The median augmentation-to-set-based radius ratio over all ten tasks is $2.03$ as a descriptive summary.}
\label{tab:tentask}
\begin{tabular}{@{}c cc cc c l@{}}
\toprule
task & aug.\ c/e & set c/e & aug.\ $q$ & set $q$ & ratio & category \\
\midrule
t0 & 0.60/0.58 & 0.30/0.34 & 0.132 & 0.048 & 2.73 & neither \\
t1 & 0.42/0.52 & 0.44/0.42 & 0.137 & 0.083 & 1.66 & neither \\
t2 & 0.84/0.90 & 0.92/0.96 & 0.147 & 0.089 & 1.65 & set-based only \\
t3 & 0.98/1.00 & 0.94/0.92 & 0.250 & 0.293 & 0.85 & both methods \\
t4 & 0.68/0.74 & 0.42/0.50 & 0.180 & 0.075 & 2.40 & neither \\
t5 & 0.96/0.84 & 0.96/0.92 & 0.163 & 0.050 & 3.25 & set-based only \\
t6 & 0.42/0.38 & 0.68/0.62 & 0.134 & 0.119 & 1.12 & neither \\
t7 & 0.78/0.66 & 0.48/0.62 & 0.430 & 0.084 & 5.13 & neither \\
t8 & 0.28/0.32 & 0.14/0.12 & 0.086 & 0.084 & 1.04 & neither \\
t9 & 0.78/0.74 & 0.60/0.62 & 0.161 & 0.053 & 3.02 & neither \\
\bottomrule
\end{tabular}
\end{table}

Table~\ref{tab:tentask} lists the full ten-task breadth that the main-paper results figure summarizes, with the matched clean and camera-extrinsic success rates of both methods. The two ratios near one, on tasks 3 and 6, indicate similar radii for the two methods under this single-seed protocol. Under a uniform best-checkpoint selection, a matched behavior evaluation of fifty clean and fifty camera-extrinsic rollouts over shared initial states with a pre-specified episode seed, and the $0.85$ minimum task-capability criterion, the augmentation baseline is capable on the one task, task 3. Set-based training is behavior-valid on tasks 2, 3, and 5, so it preserves the one baseline-capable task and recovers two further tasks, while neither method clears the criterion on the remaining tasks 0, 1, 4, 6, 7, 8, and 9. Table~\ref{tab:breadth-threshold} reports how this classification varies with the criterion, and at the looser $0.70$ gate the augmentation baseline is additionally capable on tasks 2, 5, and 9. The Task~3 row here uses this ten-task protocol with a single seed on held-out initial states 45 through 49.

\begin{figure}[t]
\centering
\includegraphics[width=\linewidth]{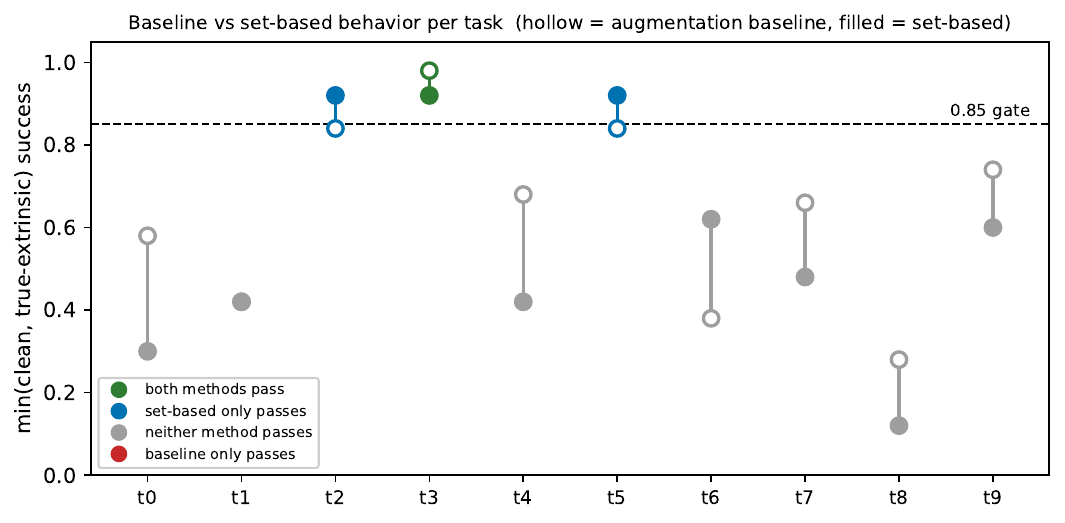}
\caption{Ten-task behavior dumbbell. For each task the hollow marker is the augmentation baseline and the filled marker is set-based training, both at the minimum of clean and camera-extrinsic success over fifty rollouts each, with the two methods connected. The dashed line is the $0.85$ gate. Color encodes which methods pass the gate, where green marks the tasks both methods pass, blue marks the tasks only set-based training passes, red marks the tasks only the baseline passes, and gray marks the tasks neither method passes. The augmentation-only category is empty at this gate. All ten tasks are shown.}
\label{fig:behavior-dumbbell}
\end{figure}

\paragraph{Cross-task behavior preservation.} We group the ten tasks by which methods pass the behavior gate at the $0.85$ minimum task-capability criterion. Both methods pass on task 3, only set-based training passes on tasks 2 and 5, no task passes only under the augmentation baseline, and neither method passes on the remaining tasks 0, 1, 4, 6, 7, 8, and 9. Figure~\ref{fig:behavior-dumbbell} shows the paired behavior of the two methods on every task, and Table~\ref{tab:breadth-threshold} reports how the classification varies with the criterion, where the baseline-only set is task 9 alone at the $0.70$ gate and is empty from the $0.75$ gate upward, so it is empty at the $0.85$ main gate.

\begin{table}[t]
\centering
\footnotesize
\setlength{\tabcolsep}{4pt}
\caption{Threshold sensitivity of the ten-task minimum task-capability classification. The criterion applies the same clean and camera-extrinsic floor to both the augmentation-only baseline and set-based training. Capable counts the tasks the baseline clears, both counts the capable tasks set-based training also clears, and baseline-only counts the capable tasks set-based training does not clear. The main paper uses the $0.85$ gate. Recomputed from the released result artifact.}
\label{tab:breadth-threshold}
\begin{tabular}{lccccc}
\toprule
gate & capable & both & base-only & both/cap. & base-only tasks \\
\midrule
0.70 & 4 & 3 & 1 & 3/4 & t9 \\
0.75 & 3 & 3 & 0 & 3/3 & none \\
0.80 & 3 & 3 & 0 & 3/3 & none \\
\textbf{0.85} & 1 & 1 & 0 & 1/1 & none \\
0.90 & 1 & 1 & 0 & 1/1 & none \\
\bottomrule
\end{tabular}
\end{table}

\section{Capacity Tiers}
\begin{figure}[t]
\centering
\includegraphics[width=0.6\linewidth]{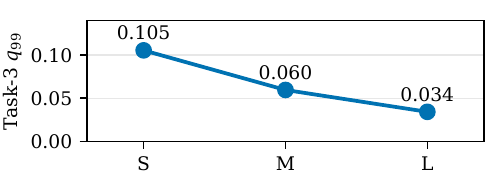}
\caption{Capacity frontier of set-based training. The Task~3 rollout-initialization-level $99\%$ conformal radius $q$ across the small, medium, and large capacity tiers of the flow matching network, each tier about $1.7$ times tighter than the one below.}
\label{fig:results-c}
\end{figure}

On the main-text sampled-camera protocol, the selected behavior-valid configuration for each capacity tier reaches a rollout-initialization-level $99\%$ radius $q$ of $\capSmallQ$ at the small tier, $\capMedQ$ at the medium tier, and $\capLargeQ$ at the large tier, each tier about $1.7$ times tighter than the one below, as Figure~\ref{fig:results-c} shows. These tiers differ in propagated width, training target, and selected checkpoint together, so the ordering describes the frontier of selected behavior-valid configurations across architecture scales and not a pure capacity effect.

\section{Terminal-Width Bound and Conformal Bridge}
Proposition~3 is stated in the main text, and we give its proof here. It bounds the physical action-deviation score $S_\theta(g,v)$ by the propagated width quantity $U_\theta(g;r)$ on the coverage event $E_r(g,v)$, gives the tail bound with the miss probability $\gamma$, and, with the sample-specific radius $r_a(v)=\lVert z_a^v-z_a^0\rVert_\infty$, yields the unconditional pointwise bound $S_\theta(g,v)\le\widetilde U_\theta(g,v)$ and the conformal ordering $q_\theta\le q_\theta^{U}$ at a shared rank $k$. The empirical support below reports the fixed-radius coverage as an empirical quantity on held-out groups, and the directly calibrated radius $q_\theta$ stays the tighter primary result.

\paragraph{Proof of Proposition~3.} With the notation of the main text, on $E_r(g,v)$ each perturbed representation $z_a^v$ lies in $\langle z_a^0,rI_m\rangle$, so by Proposition~1 the decoded action lies in $\mathcal{Z}^{\mathrm{out}}_a(r)$ and every entry obeys $\lvert f_{\theta,tj}(z_a^v,u_a)-c^{\mathrm{out}}_{a,tj}(r)\rvert\le\sum_k\lvert G^{\mathrm{out}}_{a,tjk}(r)\rvert$. The triangle inequality then gives $\lvert f_{\theta,tj}(z_a^v,u_a)-\tilde{A}_{a,tj}\rvert\le\lvert c^{\mathrm{out}}_{a,tj}(r)-\tilde{A}_{a,tj}\rvert+\sum_k\lvert G^{\mathrm{out}}_{a,tjk}(r)\rvert$, and multiplying by $s_j$ and taking the maximum over $t$, over $j\in\mathcal{C}$, and over $a\in\mathcal{A}(g)$ gives the coverage-conditional bound $S_\theta(g,v)\le U_\theta(g;r)$. The tail bound $\Pr[S_\theta(g,v)>u]\le\gamma+\Pr[U_\theta(g;r)>u]$ follows from $\{S_\theta>u\}\subseteq E_r^c\cup\{U_\theta>u\}$. Choosing $r_a(v)=\lVert z_a^v-z_a^0\rVert_\infty$ makes the inclusion hold for every sampled pair, so the same entry bound taken over $t$ and $j\in\mathcal{C}$ gives $\max_{t,\,j\in\mathcal{C}}s_j\lvert f_{\theta,tj}(z_a^v,u_a)-\tilde{A}_{a,tj}\rvert\le U_{\theta,a}(r_a(v))$ for each $a$, and taking the maximum over $a\in\mathcal{A}(g)$ gives $S_\theta(g,v)\le\max_{a}U_{\theta,a}(r_a(v))=\widetilde U_\theta(g,v)$. Pointwise ordering $S_i\le\widetilde U_i$ implies the same ordering of their $k$-th order statistics, and split conformal calibration of $\widetilde U$ gives the conformal ordering $q_\theta\le q_\theta^{U}$ with $\Pr[S_\theta(g_{\mathrm{new}},v_{\mathrm{new}})\le q_\theta^{U}]\ge1-\alpha$. $\square$

\paragraph{Empirical support.} We evaluate the bound on the same $200$ calibration and $100$ test rollout initializations and the same frozen camera poses as the main study, over the twelve Task-3 checkpoints. For each clean state the interface set $\mathcal{B}_a=\langle z_a,\epsilon I_{32}\rangle$ at the fixed training radius $\epsilon=0.240$ is propagated with the same DeepZ transformer that produces the training width, which gives the output zonotope $\mathcal{Z}^{\mathrm{out}}_a$ and hence $U_\theta(g)$. The bottleneck residual $\lVert h(o(a,v_g))-h(o(a,v_0))\rVert_\infty$ and the physical score $S_\theta$ are read directly from the sampled-camera evaluation. Table~\ref{tab:deepz-bound} reports the result. The empirical training-radius coverage on the $100$ test groups ranges from $0.76$ to $1.00$ across seeds, matching the separately reported interface-set coverage. On the covered groups \deepzViolTrain{} groups violate $S_\theta\le U_{\mathrm{train}}$, and on all groups \deepzViolAdapt{} groups violate the adaptive bound $S_\theta\le U_{\mathrm{adapt}}$, consistent with Proposition~3 and its adaptive specialization. Set-based training reduces the median propagated bound by more than two orders of magnitude at the $99\%$ level relative to every control, from a set-based median $q_{99}(U_{\mathrm{train}})$ of \deepzUsetMedian{} against roughly \deepzUctlMedian{} for the behavior-only and observational-consistency controls and \deepzUpgdMedian{} for PGD adversarial training, with the per-seed $U_{\mathrm{train}}$ and $U_{\mathrm{adapt}}$ values in Table~\ref{tab:deepz-bound}. The direct sampled score $q_{99}(S)$ stays near $0.1$ to $0.2$ for every method. Applying split conformal calibration to the adaptive bound at the same $99\%$ rank as $q_\theta$ gives the conformal quantity $q_\theta^{U}$, and $q_\theta\le q_\theta^{U}$ holds on \deepzBridgeHolds{} checkpoints, with median $q_\theta^{U}$ near \deepzQUsetMedian{} for set-based training against \deepzQUctlMedian{} for the behavior-only and observational-consistency controls and \deepzQUpgdMedian{} for PGD adversarial training. The bound is a conservative over-approximation because it adds the full-set interval half-width $\sum_k\lvert G^{\mathrm{out}}_{a,jk}\rvert$ to the center offset $\lvert c^{\mathrm{out}}_{a,j}-\tilde{A}_{a,j}\rvert$, both larger than the single sampled residual that the score $S_\theta$ measures, so it supports the mechanism that contracting the terminal width lowers a provable upper bound on the physical action deviation.
\begin{table}[t]
\centering
\small
\setlength{\tabcolsep}{4pt}
\caption{Empirical support for the coverage-conditional action bound of Proposition~3, on the same $200$ calibration and $100$ test rollout initializations and frozen camera poses as the main study, over the twelve formal checkpoints. Coverage is the empirical fraction of the $100$ test groups whose bottleneck residual stays within the fixed training radius $\epsilon=0.240$. The violation columns count groups with $S>U$ beyond $10^{-6}$, for $U_{\mathrm{train}}$ on the covered groups and the adaptive $U_{\mathrm{adapt}}$ on all groups. $U$ and $S$ are in raw action units and the unit is the rollout group. The dagger marks the behavior-invalid seed-621 PGD frontier. The direct conformal radius $q$ is the primary result and is not replaced by $U$.}
\label{tab:deepz-bound}
\resizebox{\columnwidth}{!}{%
\begin{tabular}{@{}l l c c r c r r@{}}
\toprule
Method & seed & cov. & viol.$_{\mathrm{tr}}$ & $q_{99}U_{\mathrm{tr}}$ & viol.$_{\mathrm{ad}}$ & $q_{99}U_{\mathrm{ad}}$ & $q_{99}S$ \\
\midrule
Set-based & s621 & 1.00 & 0 & 1.57 & 0 & 1.19 & 0.111 \\
 & s622 & 0.76 & 0 & 2.60 & 0 & 3.29 & 0.113 \\
 & s623 & 0.92 & 0 & 2.15 & 0 & 1.91 & 0.077 \\
Behavior-only & s621 & 1.00 & 0 & 1.01e+03 & 0 & 936.55 & 0.231 \\
 & s622 & 0.76 & 0 & 602.46 & 0 & 826.47 & 0.181 \\
 & s623 & 0.92 & 0 & 1.02e+03 & 0 & 1.22e+03 & 0.191 \\
Consistency & s621 & 1.00 & 0 & 1.01e+03 & 0 & 938.85 & 0.220 \\
 & s622 & 0.76 & 0 & 695.27 & 0 & 962.04 & 0.201 \\
 & s623 & 0.92 & 0 & 1.03e+03 & 0 & 1.23e+03 & 0.186 \\
PGD & s621$^{\dagger}$ & 1.00 & 0 & 745.21 & 0 & 686.76 & 0.173 \\
 & s622 & 0.76 & 0 & 316.70 & 0 & 421.67 & 0.138 \\
 & s623 & 0.92 & 0 & 743.39 & 0 & 888.05 & 0.114 \\
\bottomrule
\end{tabular}}
\end{table}

\section{Tightness of the Deterministic Output Enclosure}
Proposition~1 establishes that the DeepZ propagation returns an over-approximation of the action set induced by an interface set, so every action reachable from that set lies inside the returned enclosure. This section measures how tight that enclosure is on the twelve Task-3 camera checkpoints, by comparing the DeepZ terminal half-width against the strongest deviation that a projected-gradient attack attains inside the same interface set.

\paragraph{Protocol.} Each checkpoint is verified at the interface set $\mathcal{B}_\epsilon(z)=\langle z,\epsilon I_{32}\rangle$ with the shared calibrated radius $\epsilon=0.240$, which is identical across configurations because the bottleneck is frozen during every continuation. The evaluation covers \pgdNCheckpoints{} checkpoints, which are the set-based, behavior-only, observational-consistency, and PGD adversarial-training configurations across three seeds, with the seed-621 PGD checkpoint reported as a behavior-invalid frontier point, and \pgdNAnchors{} matched anchors per checkpoint drawn from the shared held-out observation bundle, which gives \pgdNEval{} checkpoint-anchor evaluations. The \pgdNAnchors{} anchors are shared across all checkpoints, so the \pgdNEval{} evaluations reuse the same \pgdNAnchors{} unique anchors, each verified under \pgdNCheckpoints{} checkpoints. The attack runs fifty restarts of one hundred steps inside the interface set, and a random baseline draws one thousand samples, both scored as the largest non-gripper deviation over the action block in raw physical action units, which is the same coordinate and scale as the DeepZ half-width.

\paragraph{Enclosure check.} Proposition~1 establishes that the enclosure contains the reachable action set. Consistent with it, every projected-gradient and random lower bound falls inside the DeepZ enclosure across all \pgdNEval{} checkpoint-anchor evaluations, so the enclosure holds empirically at every evaluation.

\paragraph{Result.} Table~\ref{tab:pgd-tightness} reports the enclosure width against the two lower bounds. For the set-based checkpoints the DeepZ enclosure is informative, with a per-configuration median DeepZ-to-PGD ratio of \pgdSetMedian{} and per-seed values from \pgdSetLo{} to \pgdSetHi{}. For each control objective the same verifier returns a far looser enclosure, with per-configuration median ratios of \pgdCtlMedian{} for behavior-only fine-tuning, \pgdConsistMedian{} for observational consistency, and \pgdPGDMedian{} for PGD adversarial training, all above \pgdControlFloor{}. The set-based enclosure widths stay within a few raw action units, whereas the three control enclosure widths are in the hundreds, so an optimized pointwise adversarial objective attains a comparable sampled radius without producing an informative deterministic enclosure. Every checkpoint-anchor pair satisfies the enclosure relations, with the DeepZ width at least the projected-gradient lower bound and at least the random lower bound on all \pgdNEval{} evaluations.

This reading is a deterministic statement about the evaluated interface sets and the output enclosure DeepZ returns for them, and the conformal radius $q$ of the main text provides the complementary probabilistic statement over camera poses. The same training signal that tightens the conformal radius also yields an informative deterministic output enclosure, whereas the control objectives leave that enclosure orders of magnitude looser under the identical verifier.

\begin{table}[t]
\centering
\small
\setlength{\tabcolsep}{4pt}
\caption{Tightness of the deterministic output enclosure on the twelve formal Task-3 camera checkpoints, the set-based, behavior-only, observational-consistency, and PGD adversarial-training configurations across three seeds. Each checkpoint is evaluated at its \pgdNAnchors{} matched anchors under the shared isotropic interface set $\mathcal{B}_\epsilon(z)=\langle z,\epsilon I_{32}\rangle$ with $\epsilon=0.240$, giving \pgdNEval{} checkpoint-anchor evaluations in raw physical action units over the six continuous non-gripper coordinates. DeepZ width is the terminal half-width, and the PGD and random columns are the strongest-attack and random lower bounds. The DeepZ/PGD columns give the median, seventy-fifth, and ninetieth percentile of the ratio over the \pgdNAnchors{} anchors. The dagger marks the behavior-invalid seed-621 PGD frontier checkpoint.}
\label{tab:pgd-tightness}
\resizebox{\columnwidth}{!}{%
\begin{tabular}{llrrrrrrr}
\toprule
config. & seed & DeepZ width & PGD & random & DeepZ/PGD med & p75 & p90 & PGD/random \\
\midrule
Set-based & s621 & 3.314 & 0.188 & 0.088 & 16.4 & 27.1 & 30.6 & 2.18 \\
Set-based & s622 & 1.740 & 0.104 & 0.040 & 17.0 & 24.3 & 31.5 & 2.46 \\
Set-based & s623 & 2.498 & 0.157 & 0.069 & 16.0 & 25.9 & 31.0 & 2.56 \\
Behavior-only & s621 & 691.088 & 0.760 & 0.302 & 932.8 & 1082.4 & 1145.4 & 2.58 \\
Behavior-only & s622 & 321.210 & 0.623 & 0.225 & 508.7 & 636.8 & 707.7 & 2.68 \\
Behavior-only & s623 & 649.899 & 0.619 & 0.292 & 1041.9 & 1186.9 & 1279.6 & 2.30 \\
Consistency & s621 & 693.287 & 0.745 & 0.308 & 934.6 & 1052.2 & 1154.3 & 2.62 \\
Consistency & s622 & 371.421 & 0.635 & 0.237 & 585.4 & 705.5 & 752.9 & 2.67 \\
Consistency & s623 & 652.502 & 0.624 & 0.297 & 1045.8 & 1199.4 & 1293.9 & 2.23 \\
PGD & s621$^{\dagger}$ & 506.025 & 0.448 & 0.197 & 1224.4 & 1353.7 & 1543.2 & 2.36 \\
PGD & s622 & 170.987 & 0.263 & 0.099 & 663.9 & 797.1 & 855.3 & 2.41 \\
PGD & s623 & 482.824 & 0.398 & 0.172 & 1238.8 & 1298.0 & 1361.2 & 2.41 \\
\bottomrule
\end{tabular}}
\end{table}

\section{Raw-1024 Bottleneck-Role Ablation}
Set-based training on the raw 1024-dimensional feature interface is trainable and reaches the requested terminal radii, but every tested pressure and checkpoint fails the closed-loop behavior gate. At the loose target, the best checkpoint reaches 0.500/0.480 clean and camera-extrinsic success, while the tighter targets reach zero success in both modes. No raw-feature contraction row is therefore eligible for action-radius evaluation. Together with the behavior-valid 32-dimensional pipeline, this result supports the learned bottleneck as a behavior-preserving set interface in the evaluated compact-policy setting.

\section{Reproducibility Configuration}
All entries are fixed before any state-level evaluation and recorded in the experiment configuration.

\begin{table}[t]
\centering
\footnotesize
\setlength{\tabcolsep}{4pt}
\caption{Central training configuration, fixed before evaluation. The same settings apply to the behavior-only fine-tuning, observational-consistency, and set-based configurations, which differ only in the training objective and share the pretrained checkpoint, cache, checkpoint schedule, and behavior gate.}
\label{tab:repro}
\begin{tabular}{@{}l p{0.62\columnwidth}@{}}
\toprule
Setting & Value \\
\midrule
Policy & flow-matching compact policy \\
Encoder / bottleneck & frozen $1024$-d encoder, learned $32$-d bottleneck (both frozen at contraction) \\
Training-set radius $\epsilon$ & $0.240$, effective generator $G_{\mathrm{eff}} = 0.240\, I_{32}$ \\
Terminal $\rho$ & $\max_j\!\big[\,s_j \sum_k |G_{jk}|\,\big]/\nu_i$, mean over anchors, in commanded action units, with $s_j$ the per-coordinate de-normalization scale and $\nu_i$ the commanded nominal action norm \\
Reported channel & $q$ over continuous coordinates, gripper tracked separately \\
Optimizer & AdamW, lr $5\times 10^{-5}$, weight decay $10^{-4}$, global grad clip $1.0$ \\
Batch size & $192$ behavior, $2$ set \\
Dual variable & rate $0.01$, clamp $[0,100]$ \\
Target radius $r_{\mathrm{target}}$ & per capacity tier (small $0.25$, medium and large $1.0$) \\
BatchNorm & running statistics frozen during contraction \\
Checkpoint schedule & epochs $40, 80, 120, 160, 200$ of $220$, select behavior-valid then min $\rho$ \\
Behavior gate & $0.85$ clean and $0.85$ camera-extrinsic, $50 + 50$ episodes \\
Camera sampling & one frozen uniform pose per initialization over $[-1,1]$ deg and mm, agentview, shared across configurations \\
Seeds & three training seeds s621, s622, s623, fixed per-episode seed formula \\
Selection budget & no reported evaluation initialization enters the training-radius calibration or checkpoint selection \\
\bottomrule
\end{tabular}
\end{table}

\section{Compute}
Training one policy configuration runs in about fourteen minutes on one GPU for $220$ epochs of $120$ steps. The sampled-camera perturbation evaluation over the $200$ calibration and $100$ test initializations, together with the behavior rollouts of fifty clean and fifty camera-extrinsic episodes per candidate, are the dominant evaluation cost. Every experiment runs on a single NVIDIA H200 GPU with $144$ GB of memory, under Rocky Linux $9.8$. The software stack is Python $3.10$, PyTorch $2.2$ with CUDA $12.1$, and NumPy $1.26$, with the LIBERO $0.1.1$ benchmark built on robosuite $1.4$ and MuJoCo $3.7$.

\end{document}